\documentclass[runningheads]{llncs}
\usepackage{graphicx}
\usepackage{booktabs}
\usepackage{multirow}
\usepackage{pifont}
\usepackage{subcaption}
\usepackage{hyperref}
\usepackage[dvipsnames]{xcolor}
\usepackage{amsfonts}
\usepackage[font=small,labelfont=bf]{caption}

\newcommand{\cmark}{\ding{51}}
\newcommand{\xmark}{\ding{55}}

\hypersetup{
    colorlinks=true,
    linkcolor=blue,
    filecolor=magenta,      
    urlcolor=cyan,
    }

\begin{document}

\title{Multiple Instance Neuroimage Transformer}

\authorrunning{A. Singla et al.}

\author{Ayush Singla\inst{1}
\and 
Qingyu Zhao\inst{1}
\and 
Daniel K. Do\inst{1}
\and Yuyin Zhou\inst{1,2} 
\and \\
Kilian M. Pohl\inst{1}
\and
Ehsan Adeli\inst{1}
}
\institute{Stanford University, Stanford, CA 94305, USA \and 
University of California Santa Cruz, Santa Cruz, CA 95064, USA \\
 \email{ayushsingla@stanford.edu, eadeli@stanford.edu}
 }

\maketitle

\begin{abstract}
For the first time, we propose using a multiple instance learning based convolution-free transformer model, called Multiple Instance Neuroimage Transformer (MINiT), for the classification of T1-weighted (T1w) MRIs. We first present several variants of transformer models adopted for neuroimages. These models extract non-overlapping 3D blocks from the input volume and perform multi-headed self-attention on a sequence of their linear projections.
MINiT, on the other hand, treats each of the non-overlapping 3D blocks of the input MRI as its own instance, splitting it further into non-overlapping 3D patches, on which multi-headed self-attention is computed. 
As a proof-of-concept,  we evaluate the efficacy of our model by training it to identify sex from T1w-MRIs of two public datasets:  Adolescent Brain Cognitive Development (ABCD) and the National Consortium on Alcohol and Neurodevelopment in Adolescence (NCANDA). 
The learned attention maps highlight voxels contributing to identifying sex differences in brain morphometry. 
The code is available at \url{https://github.com/singlaayush/MINIT}.
\end{abstract}

\section{Introduction}
Transformers, self-attention based architectures widely used in natural language processing (NLP) \cite{attention}, have recently been successfully adapted for numerous computer vision (CV) tasks, including classification \cite{vit}, detection \cite{sota-detection}, and segmentation \cite{sota-segmentation} in both images and videos \cite{vivit}. 
However, the analysis of MRIs still relies heavily on convolutional architectures \cite{ehsan-sex-cls-cnn,ehsan-cnn-rnn}. 
Noting the success of transformer models in NLP and CV, some contemporary works combine Convolutional Neural Network (CNN) encoders and decoders with transformer blocks for medical images \cite{transunet,fmri-transformer,medical-transformer}.

The transformer-based prior work on MRIs relies on CNN-encoded MRI representations as the input to the transformer blocks and involves sophisticated pre-training and fine-tuning paradigms. For instance, they train different blocks of the model \cite{medical-transformer} with differing loss objectives \cite{fmri-transformer}. Compared to CNNs, the key advantage of convolution-free self-attention based architectures, like transformers, is that the attention kernels are dynamically computed for an input region at inference \cite{attention}, whereas they are fixed after training for CNNs. This dynamic kernel computation allows for contextual information in the input regions to be taken into account, thus greatly improving the generalizability of the model. Recent developments in NLP and CV have suggested significant improvements that enable training convolution-free transformers from scratch \cite{how-to-train-your-vit,sam} with the help of data augmentation and regularization. Although some of these techniques were developed for 3D data, such as point clouds \cite{transformer-point-clouds} or videos \cite{vivit}, their usage for neuroimages is yet to be explored.

In this paper, we propose a novel Multiple Instance Neuroimage Transformer (MINiT) architecture for classification of 3D T1-weighted (T1w) MRIs. 
We first adopt the standard vision transformer models \cite{vit,vivit} to use cases involving 3D neuroimages. We refer to these new architectures as Neuroimage Transformers (NiT) and create different variants of them by incorporating various attention factorizations, similar to \cite{vivit}, and positional embedding \cite{roformer}. We then extend our models by encapsulating them with multiple instance learning (MIL) frameworks that have previously been explored for brain disease diagnosis using convolutional models \cite{carbonneau2018multiple,liu2018landmark}.
Specifically, MINiT extracts non-overlapping 3D blocks from the input volume and then treats each block as its own instance (3D neuroimage). It splits each block further into non-overlapping 3D patches, computes  multi-headed self-attention for each patch, and ultimately combines the results across all patches. As a result, MINiT aggregates feature embeddings in a hierarchical fashion, similar to \cite{pramono2019hierarchical}.

We compare MINiT with other variants of Neuroimage Transformers, recent 3D CNN models \cite{ehsan-sex-cls-cnn}, and an MIL implementation of CNNs.
Each model is evaluated on identifying sex from T1w MRIs of two large-scale adolescent brain image datasets: Adolescent Brain Cognitive Development (ABCD) \cite{abcd} and the National Consortium on Alcohol and Neurodevelopment in Adolescence (NCANDA) \cite{ncanda}. We follow the set up from prior work (e.g, in vivo neuroimaging \cite{hanggi2010sexual,sacher2013sexual} and computational learning-based methods \cite{van2018predicting,xin2019brain}) to identify morphological sex differences in brain development during childhood and adolescence. To ensure fair analysis in our study, we first preprocess the MRIs to correct for head size differences by affinely registering all MRIs to a template. All models are then trained in a supervised fashion (with no excessive pretraining/finetuning of any components). 
We finally visualize the attention maps learned by MINiT, highlighting voxels contributing to identifying sex differences.

\section{Method}
In this section, we describe our base transformer model, called Neuroimage Transformer (NiT).
Next, we present Multiple Instance NiT (MINiT).

\subsection{Neuroimage Transformer (NiT)}

\begin{figure}[t]
\includegraphics[width=\textwidth]{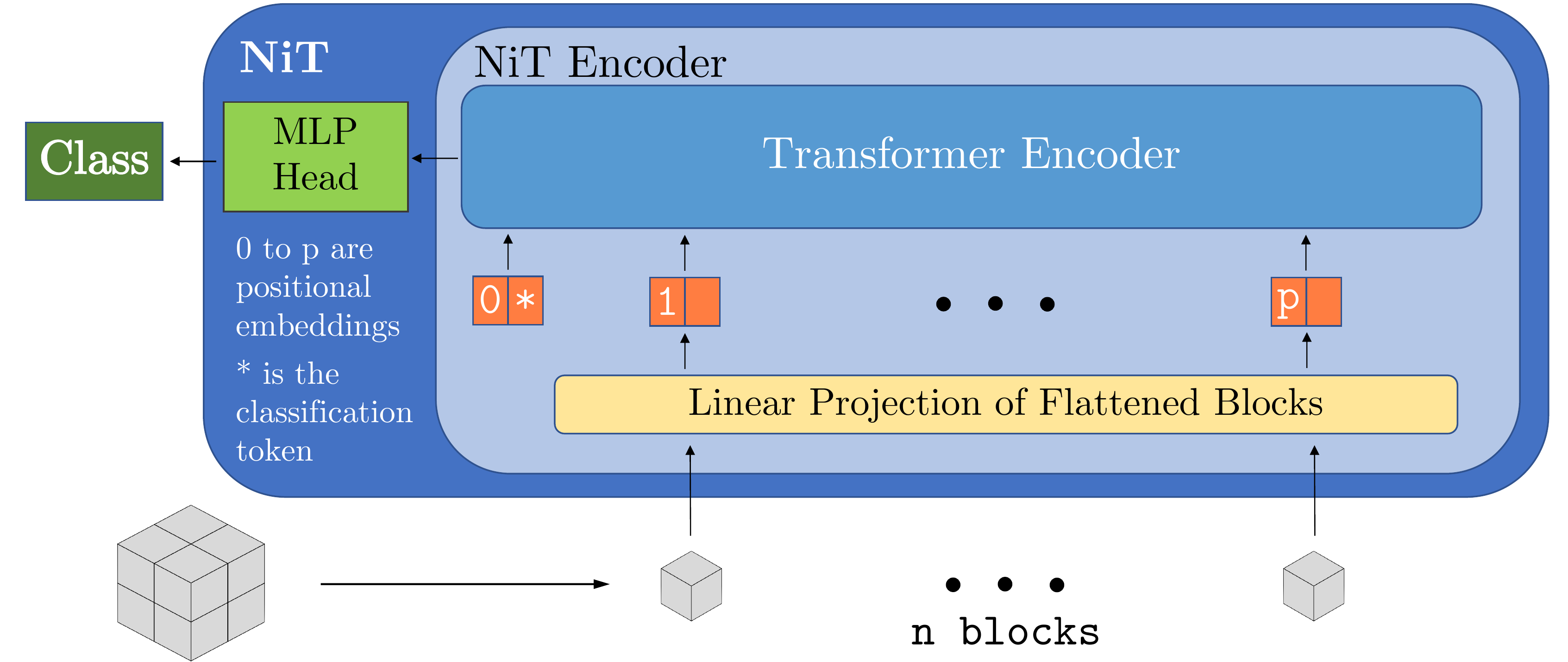} \vspace{-15pt}
\caption{Base NiT model: We split 3D neuroimages into $n$ fixed-size blocks, linearly embed each of them, add position embeddings and a class token, and feed the sequence to a standard NiT Transformer encoder.}
\label{fig:nit}
\end{figure} 
\label{subsection:nit}

For our base transformer model, we follow the overarching model design of \cite{vit}. First, we map the input neuroimage $M \in \mathbb{R}^{L \times W \times H}$ to a sequence of flattened blocks $\widetilde{z}^{n \times (B^{3})}$, where $L$, $W$, and $H$ are the length, width, and height of the input, $(B, B, B)$ is the shape of each block and $n = LWH/B^3$ is the resulting number of blocks. Similar to tubelet embeddings in \cite{vivit}, we extract non-overlapping cubiform patches from the input volume, which are subsequently flattened. Second, we project these patches to $D$ dimensions, i.e., the inner dimension of the transformer layers using a learned linear projection, generating the sequence of input patches $z^{n \times D}$. We add learned positional embeddings to retain positional information  \cite{attention} in the blocks and prepend a learned classification token \cite{bert} to their sequence serving as the input neuroimage representation.

The input sequence is then forwarded to a the transformer encoder consisting of $L$ transformer layers. Each layer contains a multi-headed self-attention (MSA) block \cite{attention} and a Multi-Layer Perceptron (MLP) block. The MLP block includes two linear projections with a Gaussian Error Gated Linear Units (GEGLU) non-linearity \cite{geglu} applied between them. Layer norm \cite{ln} is applied before and residual connections are added after every block in a transformer layer \cite{residual}. Finally, a layer norm and an MLP head consisting of a single $D \times C$ linear layer projects the classification token to $\mathbb{R}^C$, where $C$ is the number of classes (Figure \ref{fig:nit}).

\begin{figure}[t]
  \begin{minipage}[c]{0.51\textwidth}
    \includegraphics[width=\textwidth]{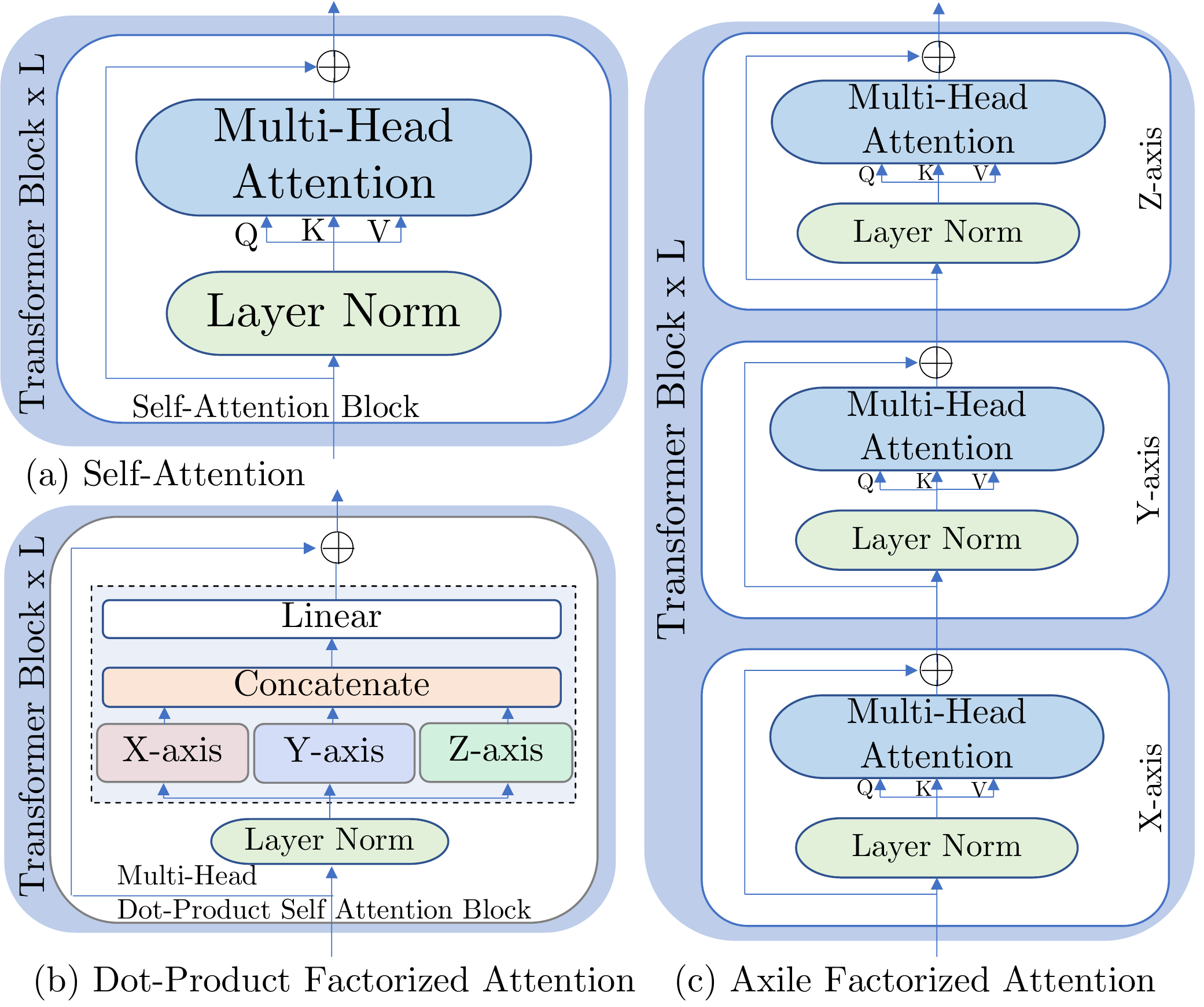}
  \end{minipage} \hfill
\begin{minipage}[c]{0.47\textwidth}
  \caption{Factorized NiT Encoders overview: (a) Vanilla NiT encoder, which uses standard MSA \cite{attention}. (b) Dot-Product Factorized NiT encoder, which factorizes MSA itself to split attention computation along all axes. (c) Axile Factorized NiT encoder, which computes factorized self-attention over all axes separately. Note that we chose the word `axile' to name this factorization to prevent confusion with commonly used `axial' plane for imaging of the brain. }
\label{fig:fact_attn}
\end{minipage}
\end{figure}

\vspace{5pt}\noindent\textbf{Factorized and Dot-Product Factorized Encoders}
\label{subsection:factorized}
Factorizing attention over input dimensions has shown effectiveness, e.g., in modeling spatio-temporal interactions in videos \cite{vivit,timesformer}, than standard self-attention. We take a similar approach and extend factorized self-attention and factorized dot-product attention by factorizing both attention and MSA over the $3$ input dimensions (Fig.~\ref{fig:fact_attn}).

For the factorized dot-product encoder (Fig.~\ref{fig:fact_attn}(b)), we factorize the MSA operation itself. We compute attention weights for each block by splitting the available attention heads into 3. Thus, a third of the attention heads are assigned to each axis dimension to compute attention by modifying the keys and values of each query in MSA to attend only over the assigned axis, as in \cite{vivit} for the temporal index. The outputs of all heads are concatenated and linearly projected to compute attention across all axes. 

In the axile factorized encoder (Fig.~\ref{fig:fact_attn}(c)), we factorize the attention operation into $3$ parts by performing MSA axially. First, we only compute MSA among all blocks along the x-axis, followed by MSA computation along the y-axis and the z-axis. We efficiently compute factorized self-attention along a single axis in the same manner that \cite{vivit} computes temporal self-attention, namely, by reshaping the flattened patches to extract the axis in question to the leading dimension.

\begin{figure}[t]
\begin{center}
\includegraphics[width=0.945\textwidth]{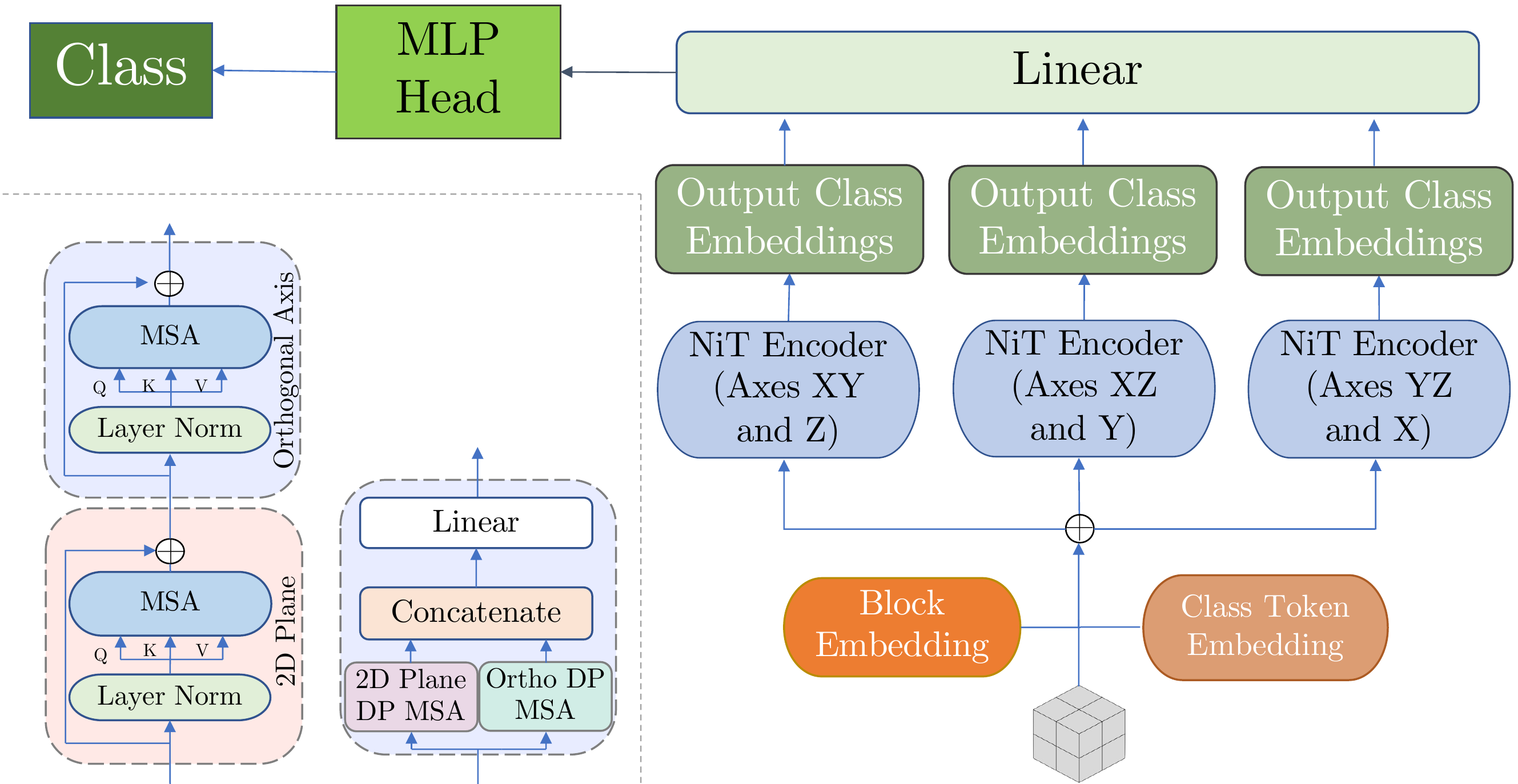}
\end{center} \vspace{-15pt}
\caption{Multi-View Factorized NiT Model. The encoders on the left show the Axile Factorized and Dot Product Factorized NiT encoders for this Multi-View approach.}
\label{fig:fact_comb_axes}
\end{figure}
\vspace{5pt}\noindent\textbf{Multi-View Factorized NiT (MVNiT)} 
The factorized self-attention methods described in \cite{vivit} differentiate the axes of the input video into spatial axes and a temporal axis. In neuroimage analysis, an analogous operation is to split a 3D neuroimage into a 2D plane combining two axes, and the orthogonal axis to the plane. We can form the 2D plane to be one of $3$ views commonly used in neuroimage analysis, namely, transverse, coronal, and sagittal. Using 2D slices of a 3D neuroimage has been found to be beneficial \cite{medical-transformer}, and thus we consider a multi-view factorized NiT  that uses factorized or dot-product factorized NiT encoders on all $3$ views (Fig.~\ref{fig:fact_comb_axes}).

We build factorized and dot-product factorized MSA blocks, which perform their respective attention operations on a combined 2D plane and the orthogonal axis. Thus, given one of the transverse, coronal, or sagittal planes with the respective orthogonal axis, the block would perform MSA treating the 2D plane as the spatial dimension and the orthogonal axis as the third dimension. 
We create three distinct encoders with these MSA blocks that consider the combined plane to be the transverse, coronal, and sagittal planes, respectively. The input sequence of patches is fed to all three encoders with distinct classification tokens, which then produce their respective class embeddings. These embeddings are concatenated and linearly projected to generate the class prediction.

\vspace{5pt}\noindent\textbf{Rotary Embeddings}
As an alternative, we modify our positional embedding to use rotary embeddings (RE) \cite{roformer}. RE has been shown to enhance prediction accuracies by incorporating explicit relative position dependency in self-attention. We adapt this method by calculating rotary embeddings along each axis, concatenating them, and then calculating self-attention as normal.

\begin{figure}[t]
\includegraphics[width=\textwidth]{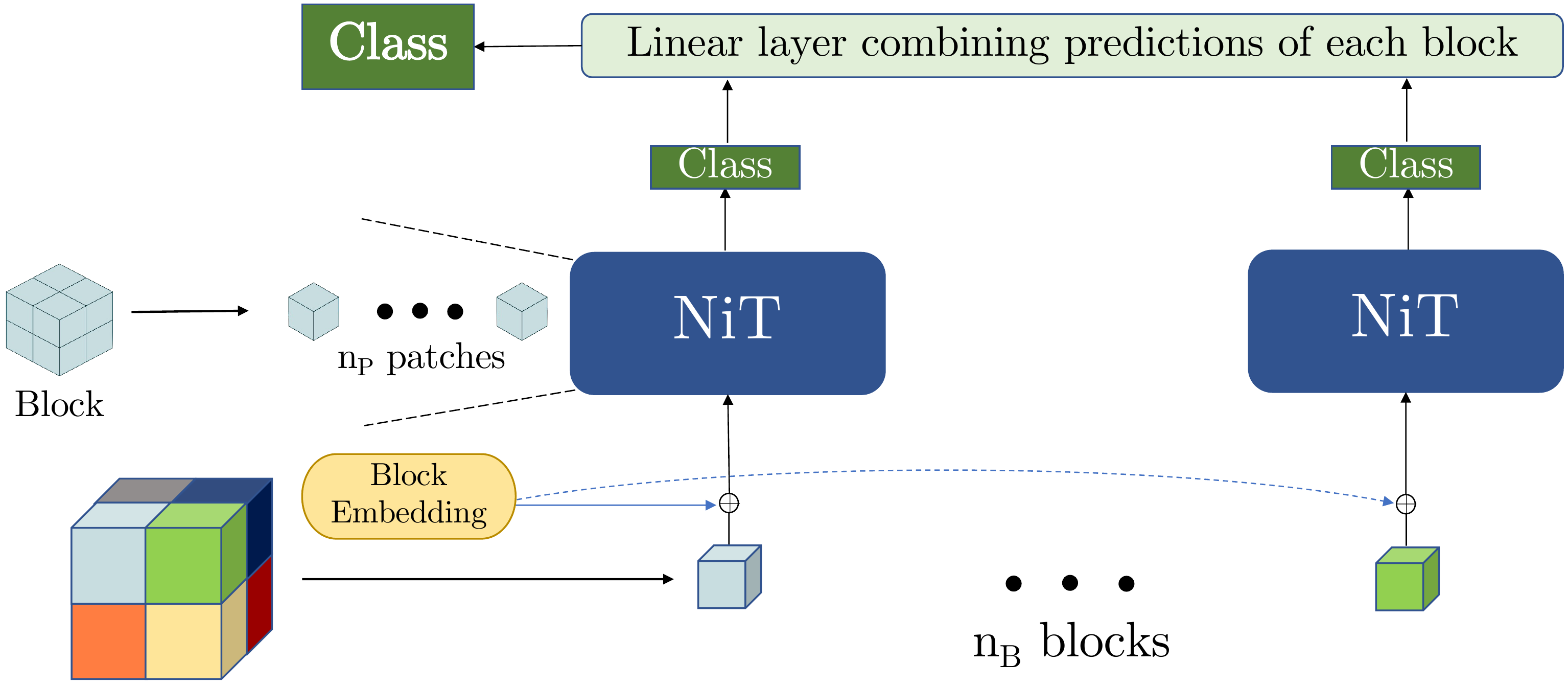} \vspace{-15pt}
\caption{MINiT: We split the 3D image into $n_B$ fixed-size blocks, add block embeddings, and feed each to a standard NiT encoder. For MINiT, the resulting sequence of predictions is concatenated and linearly projected 
to arrive at the final class predictions.}
\label{fig:mil}
\end{figure}

\subsection{Multiple Instance NiT (MINiT)}
\label{subsection:mi-nit}
Inspired by some prior MIL deep learning models applied to medical images \cite{carbonneau2018multiple,liu2018landmark}, we next develop a convolution-free transformer-based architecture inspired by the MIL paradigm, called MINiT. Given the input neuroimage, we first map it to a sequence of $n_B$ cubiform blocks. Each of these $n_B$ blocks is fed to an NiT model, in which each block is considered the input neuroimage of the model. The NiT model further maps each block to a sequence of $n_P$ smaller flattened patches. Here, each of the $n_B$ blocks is considered to be the bag of instances, while the $n_P$ flattened patches are analogous to the instances in MIL \cite{liu2018landmark} (Fig.~\ref{fig:mil}).

The sequence of patches for each block is processed similar to NiT, albeit with one modification. In addition to adding learned positional embeddings to the patches (patch embeddings), and prepending a learned classification token to their sequence, we add learned block embeddings, which retain the positional information of the block within the neuroimage to each patch. 
This is crucial to loosely emulating the benefits of non-overlapping hierarchical attention, as in \cite{pramono2019hierarchical}, because block embeddings ensure that each patch learns its position within the input neuroimage.
After this step, the patches are processed by the NiT block to produce class predictions for each block, which are concatenated and linearly projected to generate class predictions for the original input neuroimage.

\begin{figure}[t]
\includegraphics[width=\textwidth]{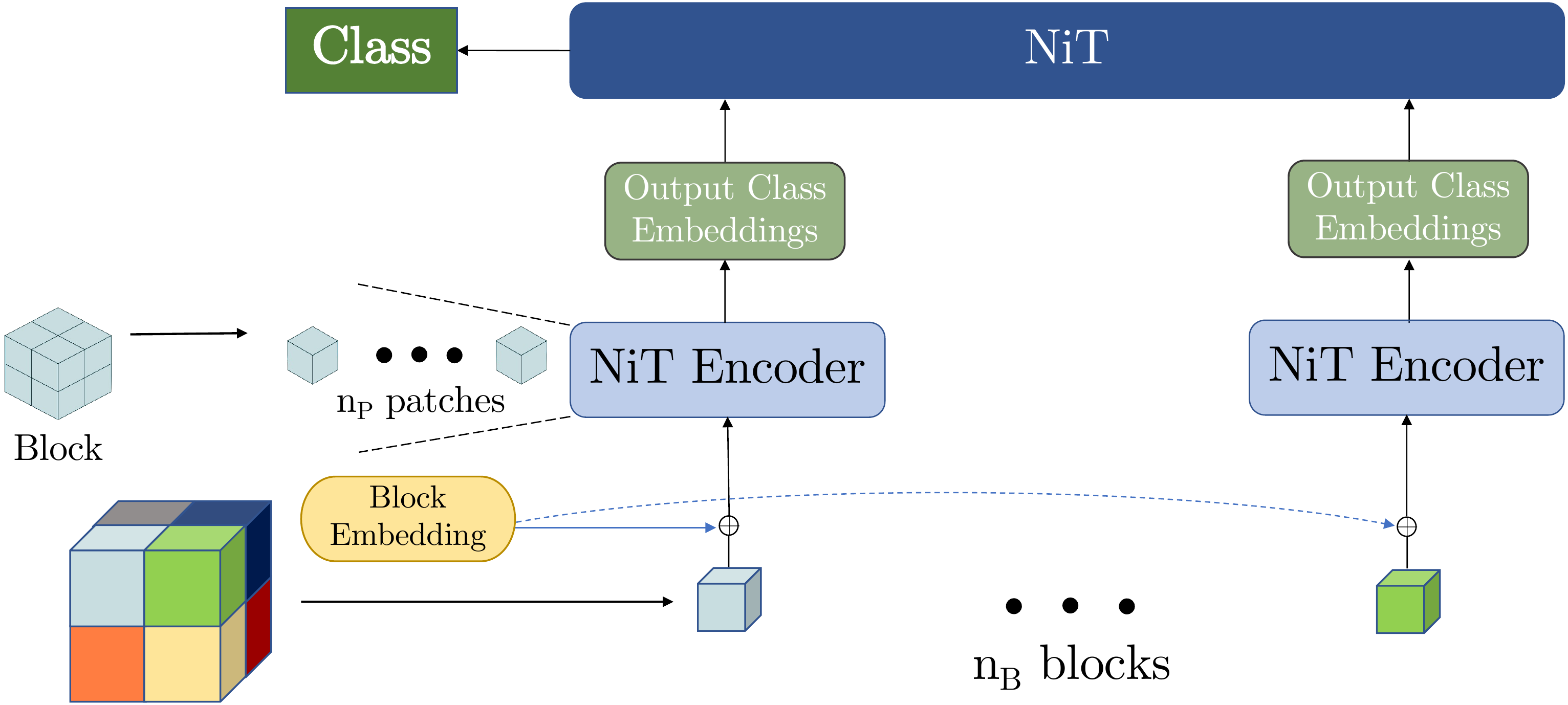}
\caption{MiGNiT overview: We split the 3D image into $n_B$ fixed-size blocks, add block embeddings, and feed each to a standard NiT encoder. The output class embeddings are input into a transformer to generate the final class predictions. }
\label{fig:mignit}
\end{figure}

\vspace{5pt}\noindent\textbf{MiGNiT: Multiple Instance Global NiT}
In this architecture, we compute global attention \cite{han2021transformer} by additionally computing self-attention on the output class embeddings produced by each block using an NiT model block. Modifying the MINiT model architecture, we change the NiT blocks to NiT encoders by stripping the final class prediction MLP head. Thus, given an individual 3D block from the input neuroimage, the NiT encoder block produces output class embeddings in $\mathbb{R}^{D}$. This sequence of output class embeddings produced from each block is fed into a complete NiT block producing class predictions (Fig.~\ref{fig:mignit}).
\section{Experiments}
As a proof-of-concept, we apply and compare MINiT with other models on identifying sex from the t1-w MRIs of  two large-scale adolescent brain image datasets: ABCD \cite{abcd} and NCANDA \cite{ncanda}.
We compare MINiT with both variants of NiTs and two CNN models, namely a 3D-CNN from \cite{ehsan-sex-cls-cnn} and an MIL-based version of a CNN, called MICNN. To identify brain regions contributing to sex classification, we visualize the attention maps learned by NiT and MINiT. 
All methods were implemented using Python 3.7.10 and its libraries (NumPy 1.15.1, Scikit-Learn 0.19.2, PyTorch 1.9.1 and TorchIO 0.18.41) on Debian GNU/Linux 10. 

\vspace{5pt}\noindent\textbf{Data}
We use $8653$ baseline T1w MRI of participants from the ABCD Study Release 2.0 (ages $10.2 \pm 0.78$ years; $52.2\%$ boys and $47.8\%$ girls).
From the NCANDA Study (release \verb|NCANDA_PUBLIC_6Y_STRUCTURAL_V01| \cite{ncanda_structural_release}), we consider the T1-w MRIs from all available visits of $808$ participants at-risk recruited between ages 12-21 years ($49.16\%$ boys and $50.84\%$ girls). Participants made $4.77$ average visits to collect $3856$ scans (age among all scans $18.77 \pm 3.14$ years).
Observed sex for all the participants across both datasets is defined as sex at birth.

In line with prior studies \cite{ehsan-confounder-cnn,ehsan-cnn-rnn}, all T1-w MRIs in the following experiments are first pre-processed by a pipeline composed of denoising, bias field correction, skull stripping, correcting for differences in head size via affine registration to a template, and re-scaling to a $64 \times 64 \times 64$ volume. 
This downsampling allows for models with smaller number of network parameters, boosting training speed. 

Post data pre-processing, our dataset combining both ABCD and NCANDA studies totals $12,506$ T1-w MRIs, which we split into training, validation, and test sets.  We use 80\% of the training split for training, 10\% for validation, and another 10\% for computing test metrics.

We further augment the training data by applying random flip and random affine transformations, adding gaussian noise, and swapping $16 \times 16 \times 16$ patches within the MRIs at random. This process increases the size of the training set by a factor of 10. 
Based on the augmented training set, we perform the training of both MINiT and all NiT and CNN based comparison models from scratch. Due to the higher proportion of MRIs provided by the ABCD study, our augmented training set has class imbalance skewed towards males. We use weighted random sampling to train all models to account for class imbalance \cite{efraimidis2006weighted}.

\vspace{5pt}\noindent\textbf{Training Strategy and Hyperparameters}
\label{subsection:training}
For all NiT based models, the $64 \times 64 \times 64$ input neuroimage volume is split into $n=8$ blocks of size $8 \times 8 \times 8$. All MIL-based models, however, use $n_B=4$ blocks of size $16 \times 16 \times 16$, which are further split into $n_P=4$ volumetric patches of size $4 \times 4 \times 4$. 
We apply dropout \cite{dropout} and weight decay \cite{decoupled-weight-decay} to both MINiT and all NiT and CNN based comparison models, as in \cite{vit,how-to-train-your-vit}.
In addition, we perform run-time data augmentations, relying on the combination of Mixup \cite{mixup} and Cutmix \cite{cutmix}. 
We train both MINiT and all NiT and CNN based comparison models using two optimizers – AdamW \cite{decoupled-weight-decay} using $\beta_1 = 0.9$ and $\beta_2 = 0.99$ and with SAM minimization \cite{sam} with Adam \cite{adam} using $\beta_1 = 0.9$ and $\beta_2 = 0.99$, and present results from the best of the two. We use a cosine decay learning rate schedule \cite{cos-lr} with gradual warmup \cite{imgnet-1hr}, and train for 200 epochs. With the exception of all MIL NiTs, which took 48 hours, all NiT models train to convergence within 24 hours on a machine with 8 NVIDIA Tesla V100 GPUs with 16GBs of memory.

We perform a search of training hyperparameters, including learning rate, dropout rates, number of warm up epochs, number of scaling epochs and their multiplier, weight decay, and probabilities to apply cutmix and mixup for both MINiT and all NiT and CNN based comparison models. For all NiT based models, we additionally perform a search for the best combination of ($L$, $N_H$, $D$, $D_{MLP}$), where $L$ is the number of transformer encoder layers, each with an MSA block of $N_H$ heads, $D$ dimension, and $D_{MLP}$ hidden MLP dimension.

\begin{table}[t]
\centering
\setlength{\tabcolsep}{3pt}
\renewcommand{\arraystretch}{0.86}
\caption{NiT model-specific hyperparameters for all NiT variants. These hyperparameters are Transformer Layers (L), $N_H$ attention heads, $D$ dimension, $D_{MLP}$ MLP dimension, Learning Rate (LR) and Weight Decay (WD).} \label{tab:hyp}
\begin{tabular}{@{}l|c|c|c|c|c|c|c|c@{}}
\toprule
\multirow{2}{*}{\textbf{Model}} & \multirow{2}{*}{\textbf{Factorization}} & \textbf{Trainable} & \multirow{2}{*}{\textbf{L}} & \multirow{2}{*}{\textbf{$N_H$}} & \multirow{2}{*}{\textbf{D}} & \multirow{2}{*}{\textbf{$D_{MLP}$}} & \multirow{2}{*}{\textbf{LR}} & \multirow{2}{*}{\textbf{WD}} \\ 
 & & \textbf{Parameters} & & & & & & \\
 \midrule
NiT & \xmark & 1.8M & 4 & 8  & 256 & 234 & $1e-4$   & 0.16 \\
NiT & Axile  & 5.1M & 6 & 8  & 256 & 64  & $1.3e-5$ & 0.05 \\
NiT & DP     & 4M & 3 & 12 & 512 & 175 & $6.5e-5$ & 0.25 \\
\hline 
MVNiT & MV \& Axile & 15M & 6 & 8 & 512 & 209 & $9e-4$ & 0.21 \\
MVNiT & MV \& DP    & 8.9M & 6 & 4 & 512 & 215 & $5e-4$ & 0.13 \\
\hline 
MiGNiT & \xmark & 8.5M & 6 & 8 & 256 & 309 & $2e-4$ & 0.3 \\
MiNiT & Axile & 3.1M  & 6 & 8 & 128 & 128 & $1e-4$ & 0.01 \\
MiNiT & DP & 3.9M & 6 & 12 & 258 & 128 & $5e-5$ & 0.24 \\
MiNiT & \xmark & 3.6M & 6 & 8 & 256 & 309 & $1e-4$ & 0.125 \\
\bottomrule
\end{tabular}
\end{table}

\vspace{5pt}\noindent\textbf{Hyperparameter Settings}
We share the NiT model-specific hyperparameters used for training in Table \ref{tab:hyp}. We also present the total number of trainable parameters for each model. The memory footprint of these models is smaller to that of most vision transformer models. Compared to small versions of contemporary vision transformer models (ViT-Small has 22.2M parameters) \cite{vit,how-to-train-your-vit}, MiNiT, has 3.6M trainable parameters. Small memory footprints make our models more appropriate for training on small datasets on smaller GPUs, thus making our work accessible to a large audience. Note that all models are trained and evaluated exactly once using these hyperparameters.

\vspace{5pt}\noindent\textbf{Evaluation Metrics}
The classification accuracy and the Area Under the Receiver Operating Characteristic (ROC) curve (AUC) of each method are computed by first binarizing the final prediction score of each participant to 0 (girl) or 1 (boy) followed by comparison to the observed sex. We additionally calculate the F-1 score (F1), sensitivity (SEN), specificity (SPE), and precision (PRC).

Apart from our NiT based comparison methods, we further compare MINiT with two contemporary CNN based models. 
First, we include the aforementioned 3D-CNN architecture from \cite{ehsan-sex-cls-cnn}, modified to accommodate $64 \times 64 \times 64$ inputs, in contrast to the original input of $64 \times 64 \times 32$. 
This CNN architecture contains $4$ convolutional blocks connected by $2 \times 2 \times 2$ 3D Max-Pooling, where each convolutional block consists of $3 \times 3 \times 3$ 3D convolution ($16$/$32$/$64$/$128$ as number of channels for the 4 blocks), Batch Normalization \cite{bn} and Re-LU \cite{relu}. 
The resulting $4,096$ dimensional features were fed into a three final linear projections of dimensions $64$, $32$, and $1$ with $tanh$, $tanh$ and $sigmoid$ activations respectively, to generate the final class predictions. 
Secondly, in line with our own MIL based approach, we create a Multiple Instance CNN (MICNN) model, which applies multiple instance learning to the inputs using the CNN framework described above. The overarching approach from Section \ref{subsection:mi-nit} remains the same, with the only difference in architecture being the replacement of the NiT Transformer with the CNN model. We equally conduct hyperparameter search for both MINiT and all NiT and CNN based comparison models.
\begin{table}[t]
\centering
\setlength{\tabcolsep}{3pt}
\renewcommand{\arraystretch}{0.86}
\caption{ACCuracy, area under the curve (AUC), F-1 score, SENsitivity, SPEcificity, and precision (PRC) for predicting sex from MRIs. Second column shows the factorization method: Axile, Dot-Product (DP), or Multi-View (MV).} \label{tab:acc}
\begin{tabular}{@{}l|c|c|c|c|c|c|c|c@{}}
\toprule
\multirow{2}{*}{\textbf{Model}} & \multirow{2}{*}{\textbf{Factorization}} & \textbf{Rotary} & \textbf{ACC} & \textbf{AUC} & \textbf{~F1~} & \textbf{SEN} & \textbf{SPE} & \textbf{PRC} \\ 
 & & \textbf{Embedding} & (\%) & (\%) & (\%) & (\%) & (\%) & (\%) \\
 \midrule
3D-CNN \cite{ehsan-sex-cls-cnn} & --- & --- & 90.4 & 96.8 & 91.5 & \textbf{94.7} & 84.9 & 88.6 \\ 
MI-CNN & --- & \xmark & 90.1 & 96.0 & 88.6 & 84.9 & 94.3 & 92.5 \\ \hline 
NiT & \xmark & \xmark & 86.4 & 93.5 & 86.7 & 87.4 & 85.4 & 86.1 \\
NiT & \xmark & \cmark & 89.1 & 95.0 & 88.9 & 87.6 & {\textbf{90.7}} & {\textbf{90.2}} \\
NiT & Axile & \xmark & 90.0 & 96.6 & 91.7 & 94.2 & 86.3 & 89.3 \\
NiT & DP & \xmark & 89.5 & 96.1 & 91.0 & 95.0 & 82.7 & 87.3 \\
\hline 
MVNiT & MV \& Axile & \xmark & 88.0 & 94.0 & 87.4 & 88.2 & 87.7 & 86.5 \\
MVNiT & MV \& DP & \xmark & 91.9 & 96.2 & 91.6 & 93.8 & 88.8 & 89.5 \\
\hline 
MIGNiT & \xmark & \cmark & 90.1 & 96.1 & 89.9 & 90.8 & 89.4 & 88.9 \\
MINiT & Axile & \xmark & 87.9 & 94.1 & 88.0 & 87.6 & 88.1 & 88.3 \\
MINiT & DP & \xmark & 90.7 & 96.2 & 89.9 & 88.3 & 92.9 & 91.6 \\
\textbf{MINiT} & \xmark & \xmark & {\textbf{92.1}} & {\textbf{97.2}} &  {\textbf{92.1}} & {94.2} & {90.0} &  {90.1} \\
\bottomrule
\end{tabular}
\end{table}

\vspace{5pt}\noindent\textbf{Results}
According to the accuracy scores in Table \ref{tab:acc}, all but one of the transformer models have better or comparable accuracy to the 3D-CNN model, with the MINiT model reporting the highest classification accuracy of $92.1\%$ as well as the highest F-1 and AUC scores of $92.1\%$ and $97.2\%$.
The MVNiT using Dot-Product Factorization also reports a comparable classification accuracy of $91.9\%$ to the MINiT model, with both models surpassing the 3D-CNN with over $1.5\%$.

MINiT's performance indicates that the use of a hierarchical attention scheme helps add positional inductive biases similar to convolutional inductive biases, while retaining the advantages of integrating information from the entire volume, even in the earlier transformer layers from self-attention \cite{vit}. These results show that our MINiT model is capable of capturing identifying characteristics between sexes, while not being insensitive to class imbalances in comparison to the 3D-CNN. The 3D-CNN is extremely sensitive to class imbalances, as evidenced by the $\approx10\%$ gap between sensitivity and specificity.
All but one of the NiT models are significantly less sensitive to class imbalances, a common problem in medical image analysis \cite{larrazabal2020imbalance}. Factorized encoders by themselves have high sensitivity to class imbalances, but using them in Multi-View or MIL settings reduces this sensitivity. MVNiT and MINiT with Axile Factorization have the lowest gaps between sensitivity and specificity, which allows them to generalize well against class imbalances, at some cost to accuracy.

\begin{figure}[t]
    \centering
    \begin{subfigure}{0.45\textwidth}
            \centering
            \includegraphics[width=0.99\textwidth]{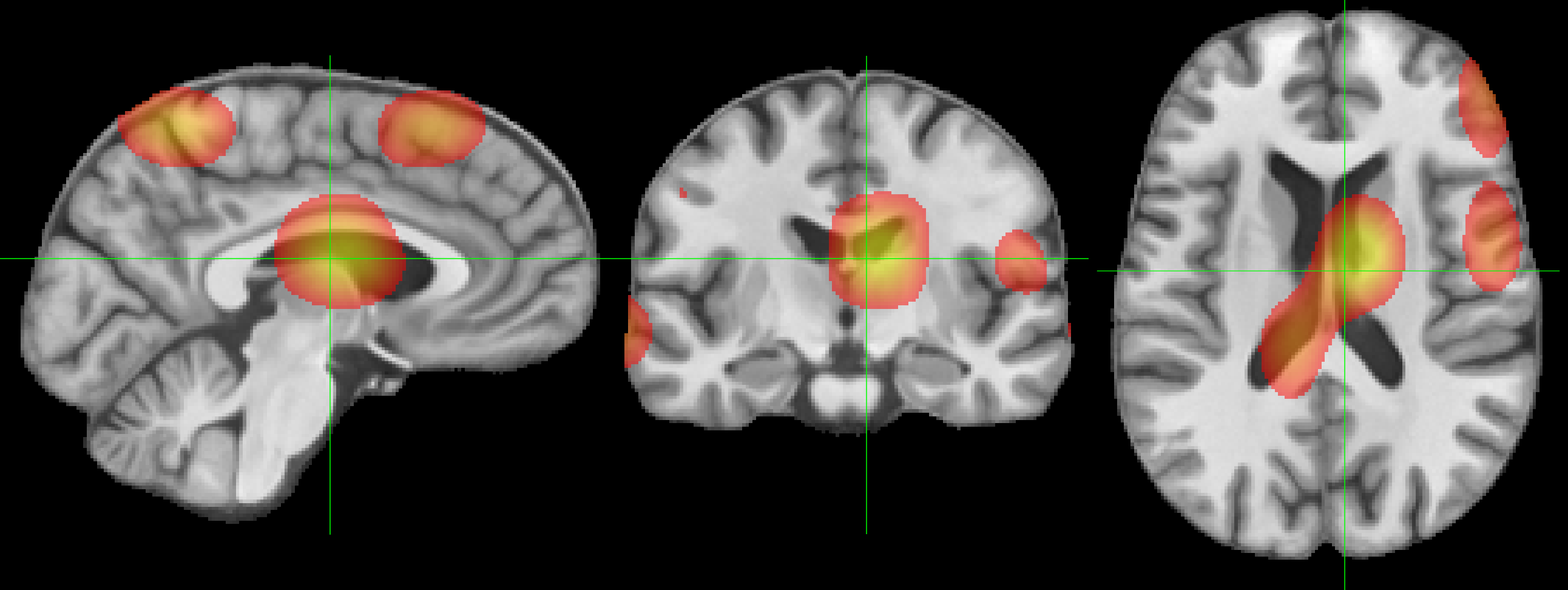}
    \caption{Base NiT Attention Map}
    \label{fig:nit-attn}
    \end{subfigure}
    \begin{subfigure}{0.45\textwidth}
            \centering
            \includegraphics[width=0.99\textwidth]{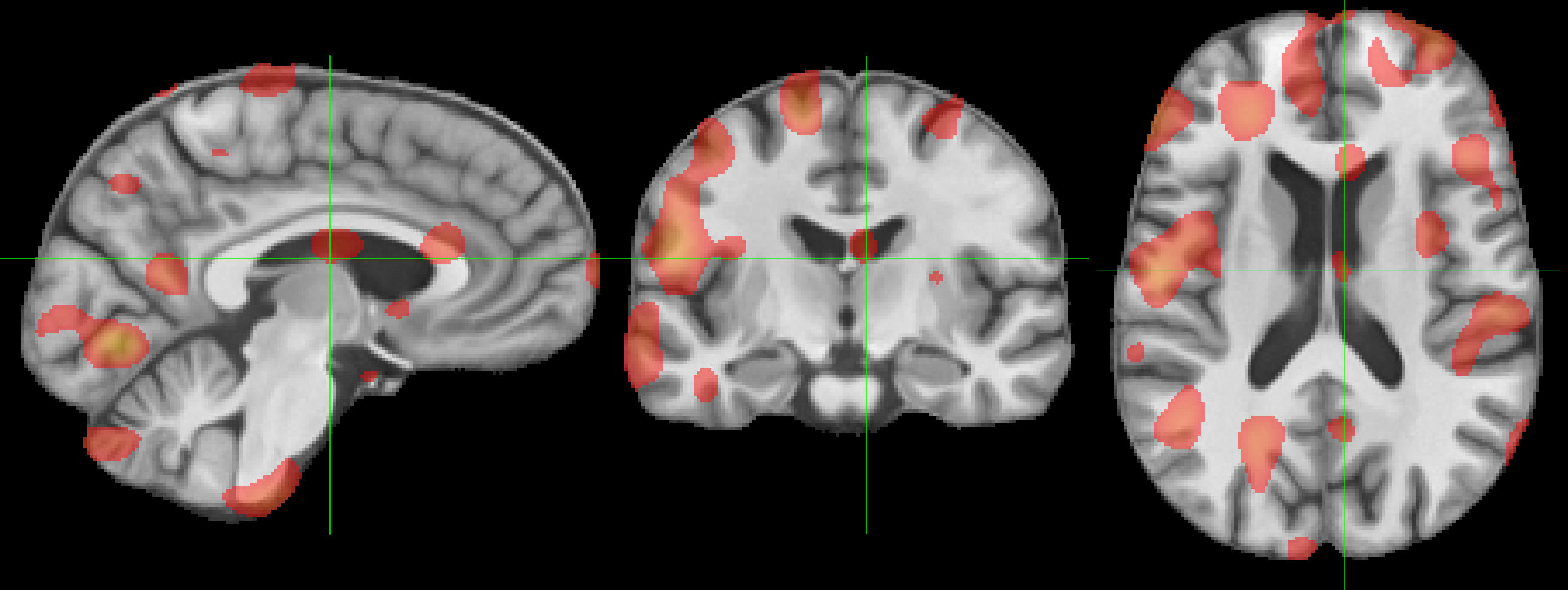}
    \caption{MINiT Attention Map}
    \end{subfigure}
    \begin{subfigure}{0.07\textwidth}
            \raisebox{0.5cm}{\includegraphics[height=2cm]{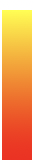}}
            \phantomcaption
    \end{subfigure} \vspace{-5pt}
    \caption{Attention maps learned by (a) NiT and (b) MINiT models. The bar shows the color-map (from red $ = 0.4$ to yellow $= 0.8$; thresholded on the lower bound for clarity). }
    \label{fig:mil-attn}
\end{figure}
We compute the attention maps for the base NiT from the output token to the input space using Attention Rollout \cite{abnar2020quantifying}. For MINiT, we use Attention Rollout to calculate attention weights for each patch in a block, which we concatenate and then average to build attention weights for a block. We proceed to use Attention Rollout using the block attention weights to compute the final attention maps.
From Fig.~\ref{fig:mil-attn}, we observe that MINiT attends between numerous different voxels in the neuroimage (due to MIL nature), in contrast to the focused attention between fewer, but larger, voxels by the base NiT. Considering existing documented evidences \cite{kaczkurkin2019sex} that sex differences in youth are widespread in the brain and the significant difference in accuracy between the two models, it is evident that MINiT is able to better generalize by capturing features spread all around the brain.

\section{Conclusion}
In this paper, we propose Multiple Instance Neuroimage Transformer (MINiT), a multiple instance learning based convolution-free transformer model for classification of 3D T1w MRIs. 
They consider the entire 3D volume and train end-to-end in a supervised fashion, with no excessive pre-training or fine-tuning required.
As a proof-of-concept, we train MINiT on identifying sex from T1w MRIs and obtain state-of-the-art results.
The visualization of the attention map learned by our MiNiT model demonstrates its ability to sensitively capture identifying differing morphological characteristics between sexes, while not being insensitive to class imbalances.
Further extensions could investigate the transfer learning capabilities of MINiT by fine-tuning on small-sized datasets for different tasks.

\vspace{5pt}\noindent\textbf{Acknowledgements}
This work was partially supported by the NIH grants AA021697 and AA028840, and the Stanford Institute for Human-centered Artificial Intelligence (HAI) Google Cloud Credits (GCP) credits.

\bibliographystyle{splncs04}
\bibliography{refs}

\end{document}